\documentclass[journal]{IEEEtran}
\usepackage{amssymb}
\usepackage{amsmath}
\usepackage{float}
\usepackage{graphics}
\usepackage{graphicx}

\usepackage[utf8]{inputenc}

\usepackage{colortbl}
\usepackage{longtable}
\usepackage{multirow}
\usepackage{verbatim}
\usepackage{epstopdf}
\usepackage[section]{placeins}
\usepackage{ifpdf}
\usepackage{cite}
\usepackage{algorithmic}
\ifCLASSOPTIONcompsoc
\usepackage[caption=false,font=normalsize,labelfont=sf,textfont=sf]{subfig}
\else
\usepackage[caption=false,font=footnotesize]{subfig}
\fi
\usepackage{url}

\begin{document}

	
%
\title{Segmentation of skin lesions based on fuzzy classification of pixels and histogram thresholding}
%
%
%

\author{Jose Luis Garcia-Arroyo,
        Begonya Garcia-Zapirain
\thanks{J.L. Garcia-Arroyo was with the Deustotech-LIFE Unit (eVIDA), University of Deusto Avda. Universidades, 24. 48007 Bilbao, Spain (e-mail: jlgarcia@deusto.es).}
\thanks{B. Garcia-Zapirain was with the Deustotech-LIFE Unit (eVIDA), University of Deusto Avda. Universidades, 24. 48007 Bilbao, Spain (e-mail: mbgarciazapi@deusto.es).}
}

%
%

\markboth{}
{Garcia-Arroyo and Garcia-Zapirain Segmentation of skin lesions based on fuzzy classification of pixels and histogram thresholding}
%



\maketitle

\begin{abstract}
This paper proposes an innovative method for segmentation of skin lesions in dermoscopy images developed by the authors, based on fuzzy classification of pixels and histogram thresholding.
\end{abstract}

\begin{IEEEkeywords}
Pattern recognition, Image processing, Machine learning, Segmentation, Border detection.
\end{IEEEkeywords}

%
\IEEEpeerreviewmaketitle

\section{Introduction}
\label{Introduction}
%
%
%
%

\IEEEPARstart{A}{utomated} segmentation of skin lesions in dermoscopy images is currently a challenging problem \cite{Celebi2015}. This paper proposes an innovative method to address this problem developed by the authors. It has been structured as follows. Firstly, in this introduction, on the one hand the segmentation problem is described and, on the other, the evaluation criteria used (image database, ground truths and metrics) are shown. Secondly, the system design is presented. Thirdly, the results and the discussion are shown.





\subsection{Problems with segmentation of skin lesions in dermoscopy images}
\label{Problematica}


Automated segmentation of a skin lesion is a complex issue, as the possible casuistry that can appear in the images is very diverse. The main problems that can de found in the image which make segmentation difficult are as follows:  


1. Presence of hair; 2. Other artifacts such as electronic letters, rulers, ink and color charts, etc.; 3. Dark rectangular or circular marks around it (a consequence of shadow); 4. Flashes; 5. Lighting problems: apart from the problem with dark marks and flashes that have already been mentioned, in some cases one part of the image turns out to be darker than another (a common cases is that the part of the skin beside the circular marks is often darker as it is less brightly lit, and some images also turn out to be darker than others; 6. As a  result of the oil used to acquire many images, there may be distortion problems and bubbles; 7. Presence of blood vessels; 8. Presence of regression areas and blue-whitish veil --in many cases these structures have greater intensity than the skin surrounding the lesion; 9. Hypopigmentation areas which are confused with skin; 10. Many colors in the image; 11. Different lesions within the same image; 12. Presence of inflamed area around the lesion; 13. Low contrast of the lesion in relation to the skin; 14. Hardly any skin is visible on occasions, and there is even no skin at all in some images; 15. What in some images has skin color and texture, in others is part of the lesion; 16. Skin color and texture is different among individuals.


Moreover and in terms of ground truths, as shown in the studies \cite{Binder1995,Joel2002,Peruch2014}, a great inter and even intra-observer variability among dermatology experts exists in producing segmentations. Such diverse casuistry that can be presented, the main problems of which have been shown above, together with, above all, subjectivity, are the causes of this variability. 


These problems clearly make segmentation difficult from the standpoint of image digital processing and, although several good methods have been described in recent years, this remains a challenging problem and the proposed method addresses them.


\subsection{Evaluation of the method: image database, ground truths and metrics}
\label{Evaluation of the method}


In order to evaluate the method, the image database, ground truths and metrics proposed by the 2016 \cite{Gutman2016} --that allowed the largest comparative study to be carried out so far among methods for segmentation of skin lesions in dermoscopy images-- and 2017 \cite{ChallengeISBI2017}, presented at the ISBI (International Symposium on Biomedical Imaging) Challenges, hosted by the ISIC (International Skin Imaging Collaboration), in which the authors participated, were adopted. 


The ``Training data set'' and ``Test data set'' were used with 2000 and 600 images respectively in the 2017 ISBI Challenge and 900 and 379 images respectively in the 2016 ISBI Challenge, selected from the ISIC archive \cite{ISIC2016b}. Although these were all taken from the same archive, they came from different sources --the concept of ``source'' is very important when analyzing a data set, since each source normally corresponds to different dermatologists, dermoscopes, techniques and therefore the features of the images-- with resolutions ranging from ${768\times576}$ to ${6748\times4499}$, and a ratio width/height from ${1.33}$ --in most cases-- to ${1.5}$. The ground truths are performed by internationally-renown dermatologist experts and all the methods are evaluated and compared using the same state-of-the-art metrics, which will be shown in \ref{Results and discussion}.

\section{System Design}
\label{System Design}

\subsection{High level view of the System Design}
\label{High level view}



Following an exhaustive review of the state-of-the-art and some preliminary tests by the authors, the following was able to be ascertained: 1. Pixel classification from color and texture features works well in detecting disturbing artifacts; 2. In the great majority of images, histogram thresholding methods for segmentation work well if the rest of the image is taken following suitable detection of disturbing artifacts; 3. Furthermore, in many cases, pixel classification from color and texture features even work fairly well for segmentation of the lesion. 


As regards classification of pixels of this type --as commented, based on color and texture features-- if the different cases of images are examined, it can be ascertained that the pixel features of color and texture vary hugely in the different images for the ``lesion'', ``skin'' and ``other'' categories, meaning that a pixel with certain features corresponds to a category in one place and that another pixel with the same features corresponds to another category in another place. Thus, a hard classification of pixels does not adapt well to the problem. Therefore, the most suitable approach is to carry out a soft or fuzzy classification --undertaken here-– that also enables a subsequent parametrization regarding possible probability values that can be taken for extraction of  ${\alpha-cuts}$ from the corresponding fuzzy sets.

\begin{figure}[hbt]
	\centering
	\centerline{
		\includegraphics[width=1\columnwidth]{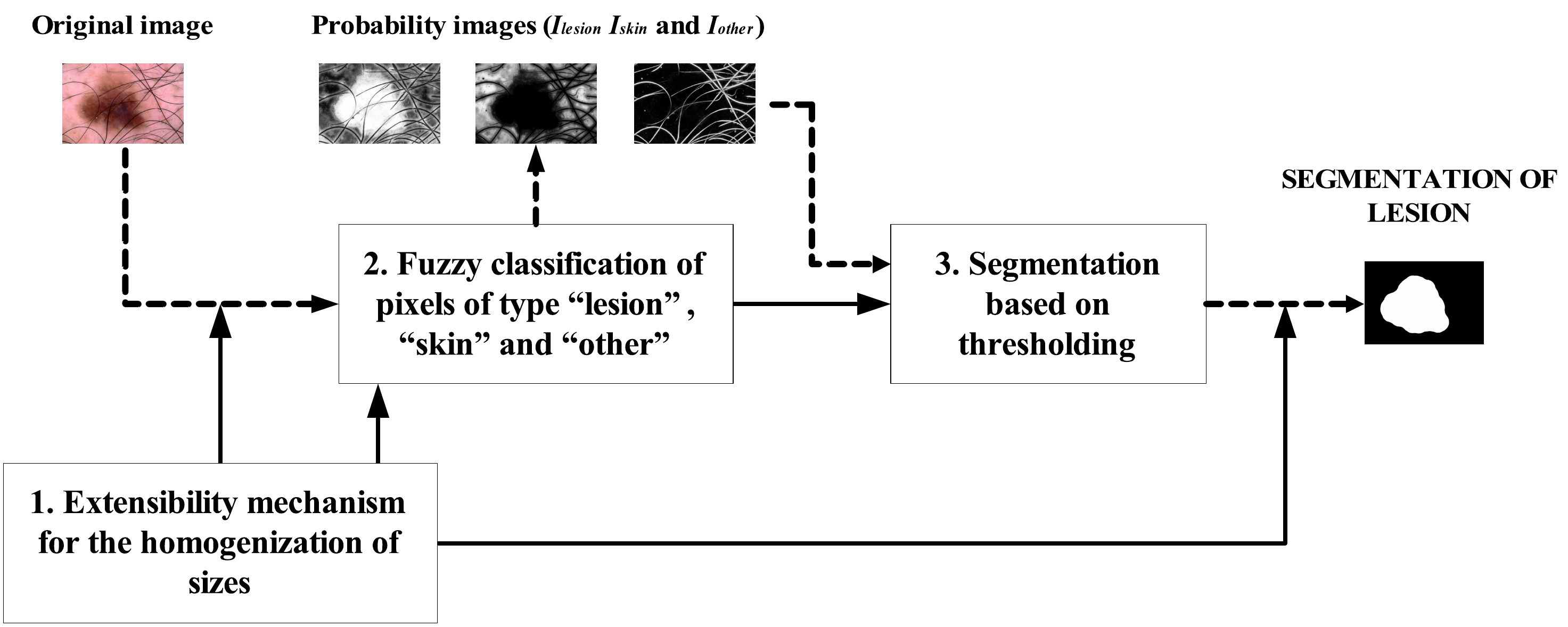}
	}
	\caption{High Level View of the Proposed System.}
	\label{fig_high_level_view}
\end{figure}

Below is shown the high level view of the system design, which is explained in detail in \ref{Scalability a imagenes tomadas de multiples fuentes}, \ref{Fuzzy classification of pixels of type ``lesion'', ``skin'' and ``other''} and \ref{Segmentation based on thresholding}. As can be seen in Fig. \ref{fig_high_level_view}, it consists of the following modules: 1. \textbf{Extensibility mechanism for the homogenization of sizes}, in which the original images are resized in such a way that they may all end up having similar sizes as a way of starting Module 2, with the reverse process being carried out following execution of Module 3 in order for the size of the mask resulting from the sizes of the original image to be re-established; 2. \textbf{Fuzzy classification of pixels of type ``lesion'', ``skin'' and ``other''}, in which a fuzzy classification of pixels is made via supervised machine learning that enables the three corresponding fuzzy sets to be created ${\mu_{lesion}}$, ${\mu_{skin}}$ and ${\mu_{other}}$ and consequently the three probability images that map them out to be generated ${I_{lesion}}$, ${I_{skin}}$ and ${I_{other}}$; 3. \textbf{Segmentation based on thresholding}, in which different operations are carried out using these probability images to detect any disturbing artifacts and, therefore, segmentation of the pixels from the lesion and skin, and then applying a histogram thresholding method for segmentation of the lesion and a subsequent post-processing. 

Finally, It is important to note that this method was designed with a view to ensuring the most accurate possible segmentation, albeit always giving priority to sensitivity over specificity, i.e. trying wherever possible to ensure that there is a minimum number of lesion pixels outside the mask resulting from the segmentation process.

\subsection{Module 1. Extensibility mechanism for the homogenization of sizes}
\label{Scalability a imagenes tomadas de multiples fuentes}


The extensibility mechanism for the homogenization of sizes is put into practice in this module. To do this, the images are downsized before starting Module 2 to a fixed width of 768 --the minimum width size of the data sets, obviously preserving the width/height proportions-- so that these downsized images, with similar sizes, constitute the input for Module 2. The reverse process is performed after Module 3 by upsizing the segmentation mask output of Module 3, so that the result is in accordance with the dimensions of the original image. The images are resized (down and upsizing) using bicubic interpolation, a fast method that reasonably maintains the image properties when downscaling \cite{Lehmann1999}. This extensibility mechanism allows the method to be fast by performing the calculations on images of minimum sizes and is also robust against the presence of images of different sizes, since the operations can be carried out in the same way on such images, critical in the fuzzy classification of pixels --which uses texture features-- and in the choice of threshold values.


\subsection{Module 2. Fuzzy classification of pixels of type ``lesion'', ``skin'' and ``other''}
\label{Fuzzy classification of pixels of type ``lesion'', ``skin'' and ``other''}

\begin{figure}[hbt]
	\centering
	\centerline{
		\includegraphics[width=1\columnwidth]{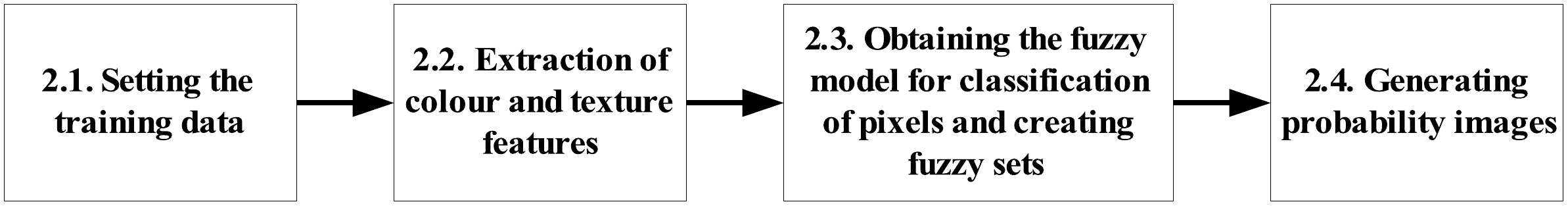}
	}
	\caption{Phases of the fuzzy classification of pixels of type ``lesion'', ``skin'' and ``other'' module.}
	\label{fig_deteccion_pixels_SEGMENTACION}
\end{figure}

In this module, the dermoscopy image pixels are fuzzy classified into ``lesion'', ``skin'' and ``other'' categories by means of a supervised machine learning process, which enables the corresponding three probability images to be generated. As can be seen in Fig. \ref{fig_deteccion_pixels_SEGMENTACION}, this module comprises 4 phases. Firstly, pixel samples are taken and labeled into the three different categories. Secondly, a set of features that are suitable for discrimination is extracted. Thirdly, this enables a fuzzy classifier to be used in order to generate a fuzzy classification model, assigning to every pixel ${(x,y)}$ a probability value belonging to each category. This in turn enables three fuzzy sets ${\mu_{lesion}}$, ${\mu_{skin}}$ and ${\mu_{other}}$ to be created on the set of pixels. Lastly, the three probability images ${I_{lesion}}$, ${I_{skin}}$ and ${I_{other}}$ are constructed from these fuzzy sets.



\subsubsection{2.1. Setting the training data}
\label{toma_muestras_pixels}


Samples of images are selected of 40 images from the ``Training data set'' from the 2016 ISBI Challenge \cite{Gutman2016}, already rescaled to a width of ${768}$, which are then labeled into the categories  ``lesion'', ``skin'' and ``other'', corresponding to the pixels in the lesion, those on the skin and the rest. The images are selected in such a way as to produce examples of each source, with different features, and an attempt is made to ensure that the amount of pixels sampled is reasonably balanced and also, as far as possible, with regard to the different casuistry existing in each category, with samples being taken in the ``other'' case corresponding to the different artifacts present in the image such as hair, rulers, flashes and bubbles.

\subsubsection{2.2. Extraction of colour and texture features}
\label{Extraction of features}



A set of color and texture features is extracted in order to characterize the pixels, with a view to discriminating between categories as far as possible. 159 features are extracted in total, of the following two types: 

\paragraph{Color features}


16 color features are extracted corresponding to the gray value and to the values of the different channels RGB, rgb (normalized RGB), HSV, CIEXYZ, CIELab and CIELuv color spaces \cite{Gonzalez2008}.

\paragraph{Texture features}


143 texture features are extracted from the image converted to gray using the formula ${I_{G}(x,y)=\frac{1}{3}I_{red}(x,y) + \frac{1}{3}I_{green}(x,y) + \frac{1}{3}I_{blue}(x,y)}$. Values are extracted  both from the gray image and from the blurred images resulting from application of a Gaussian filter bank, using the formula \cite{Gonzalez2008}: ${G_{\sigma}(x,y)=\frac{1}{2\pi\sigma^2}e^{-\frac{x^2+y^2}{2\sigma^2}}}$, with ${\sigma}$ values of the form ${\sigma=2^m}$, with ${m=0,1,2,\dots,m_{max}}$ and ${m_{max}=4}$.


- \emph{Pixel values:}
5 features are extracted corresponding to the value of each pixel: 1 for each ${\sigma}$ (the one corresponding to the gray image is considered to be a color feature).

- \emph{Sobel filter:}
6 features are extracted, corresponding to the gradient at each pixel \cite{Gonzalez2008}: 1 + 1 for each ${\sigma}$.

- \emph{Difference of Gaussian (``DoG''):}
10 features are extracted for the different pairs of values $(\sigma_{i},\sigma_{j})$, such as ${i>j}$ and ${\sigma_{m}=2^m}$, with ${m=0,1,\dots, m_{max}}$, and the different \cite{Petrou2006} ${DoG_{\sigma_{i}\sigma_{j}}(x,y)=G_{\sigma_{i}}(x,y)-G_{\sigma_{j}}(x,y)}$ are applied: corresponding to the different combinations of $(\sigma_{i},\sigma_{j})$.

- \emph{Laplacian filter:}
5 features are extracted, calculating the Laplacian \cite{Petrou2006} at each pixel: 1 for each ${\sigma}$.

- \emph{Hessian matrix:}
48 features are extracted, firstly obtaining the Hessian matrix at each pixel and then calculating 8 different features from it \cite{Petrou2006}: 8 + 8 for each ${\sigma}$.

- \emph{Texture statistics:}
25 features are extracted, with different statistics within a radius of ${\sigma}$ from each pixel (mean, variance, median, minimum and maximum) being calculated \cite{Petrou2006}: 5 for each ${\sigma}$.

- \emph{Gabor filters:}
44 features are extracted, with different Gabor filters \cite{Petrou2006} being calculated at each pixel in the gray image corresponding to different values of the parameters ${\lambda}$, ${\theta}$, ${\psi}$, ${\sigma}$ and ${\gamma}$.

\subsubsection{2.3. Obtaining the fuzzy model for classification of pixels and creating fuzzy sets}
\label{Obtaining the clasification model}


A fuzzy classifier is used to generate a fuzzy classification model from the values obtained from the labeled pixels of the different categories from the extraction of color and texture features. This model enables the probabilities of belonging to the ``lesion'', ``skin'' and ``other'' categories  to be obtained for each image pixel ${I}$. The classifier used and the results obtained are explained in \ref{Results Fuzzy classification of pixels of lesion, skin and other}.

If we consider the image ${I}$ to be of ${w\times{h}}$ in size and we define the set of pixels ${X=[0,w-1]\times[0,h-1]}$, we can then define three fuzzy sets of ${X}$ as follows. ${{\mu_{lesion}:X\xrightarrow{}[0,1]}}$, ${{\mu_{skin}:X\xrightarrow{}[0,1]}}$ and ${{\mu_{other}:X\xrightarrow{}[0,1]}}$, such that for each pixel ${(x,y)\in X}$ the values ${\mu_{lesion}(x,y)}$, ${\mu_{skin}(x,y)}$ and ${\mu_{other}(x,y)}$ are the probabilities given by the fuzzy classification. These fuzzy sets meet two criteria: firstly, they are not null, i.e. that ${\sum_{(x,y)\in X}\mu_{lesion}(x,y)>0}$, ${\sum_{(x,y)\in X}\mu_{skin}(x,y)>0}$ and ${\sum_{(x,y)\in X}\mu_{other}(x,y)>0}$) are met, and secondly, that ${\forall(x,y)\in X}$, ${\mu_{lesion}(x,y)+\mu_{skin}(x,y)+\mu_{other}(x,y)=1}$ is met. Thus, we can consider the family of fuzzy sets ${\{\mu_{lesion},\mu_{skin},\mu_{other}\}}$ to be a fuzzy partition of ${X}$.

\subsubsection{2.4. Generating probability images}
\label{Generation of probability images}

From the fuzzy sets ${\mu_{lesion}}$, ${\mu_{skin}}$ and ${\mu_{other}}$ three grey probability images are generated ${I_{lesion}}$, ${I_{skin}}$ and ${I_{other}}$ of ${w\times{h}}$ in size and a greyscale in ${[0,255]}$. These are defined as follows: ${\forall(x,y)\in X}$, ${I_{lesion}(x,y)=255.\mu_{lesion}(x,y)}$, ${I_{skin}(x,y)=255.\mu_{skin}(x,y)}$ and ${I_{other}(x,y)=255.\mu_{other}}$\\
${(x,y)}$. Evidently, ${\forall(x,y)\in X}$, ${I_{lesion}(x,y)+I_{skin}(x,y)+}$\\
${I_{other}(x,y)=255}$. 

\begin{figure}[hbt]
	\centering
	\centerline{
		\includegraphics[width=0.2\columnwidth]{ISIC_0014876-eps-converted-to.pdf}
		\includegraphics[width=0.2\columnwidth]{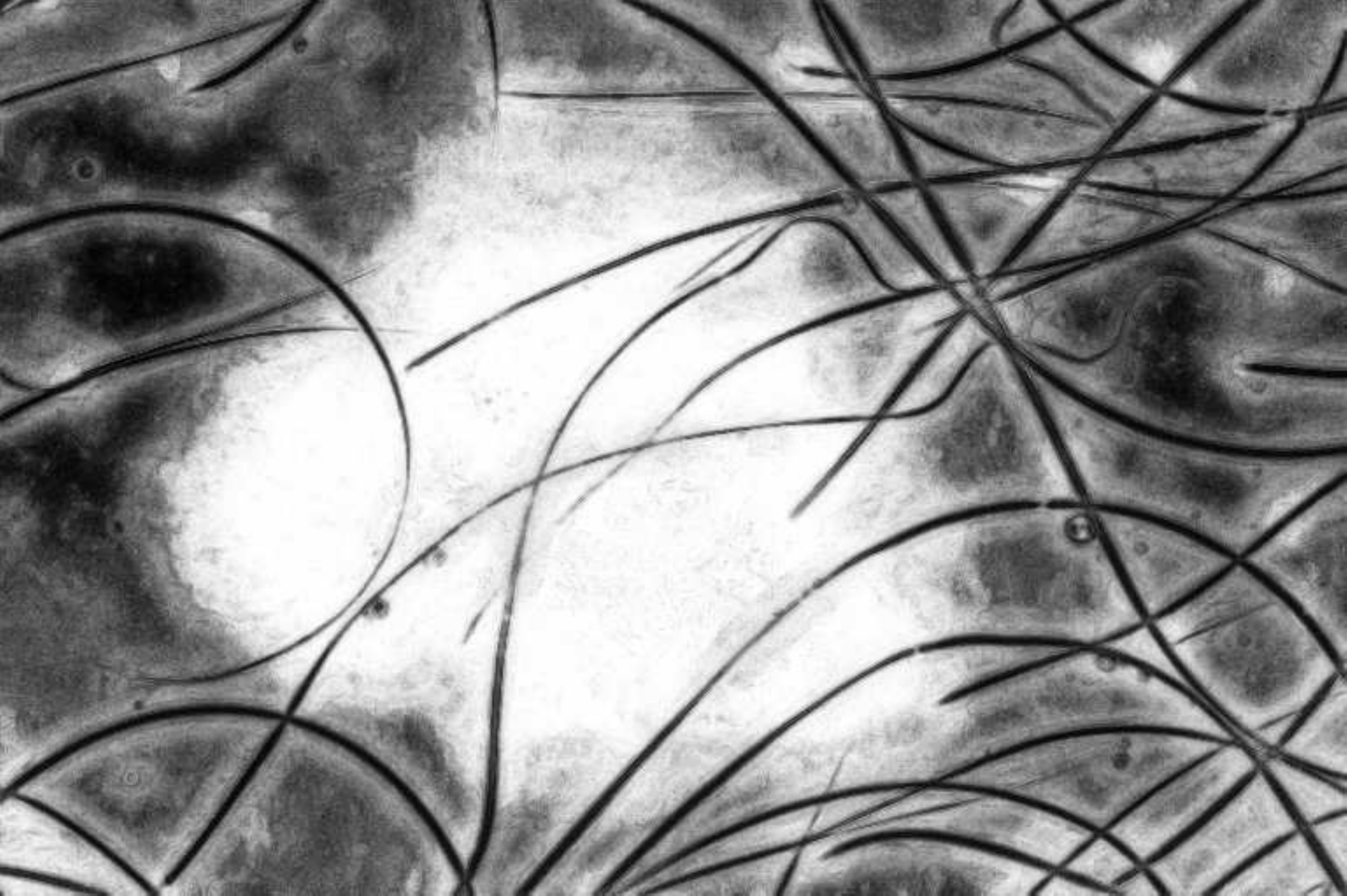}
		\includegraphics[width=0.2\columnwidth]{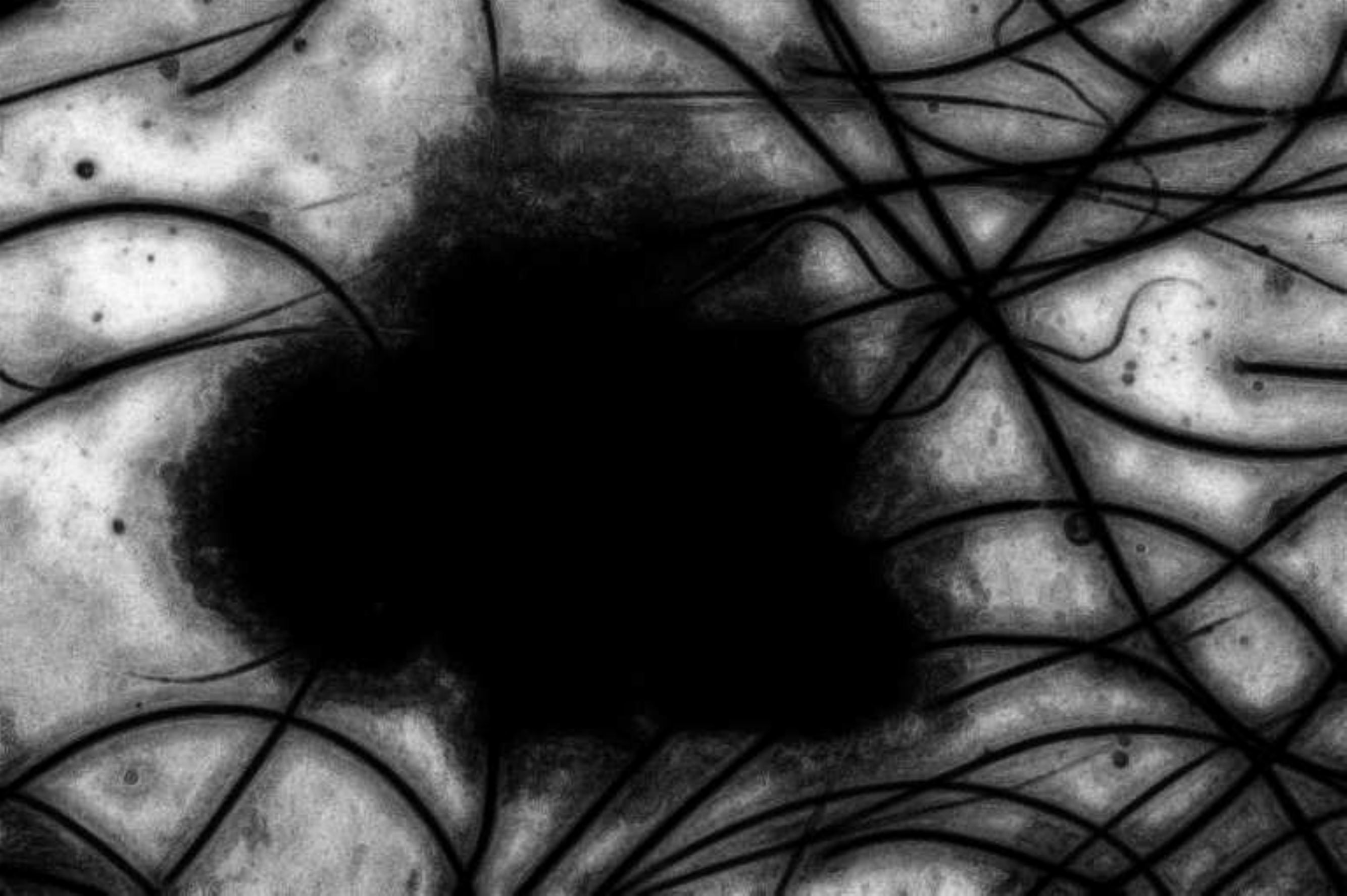}
		\includegraphics[width=0.2\columnwidth]{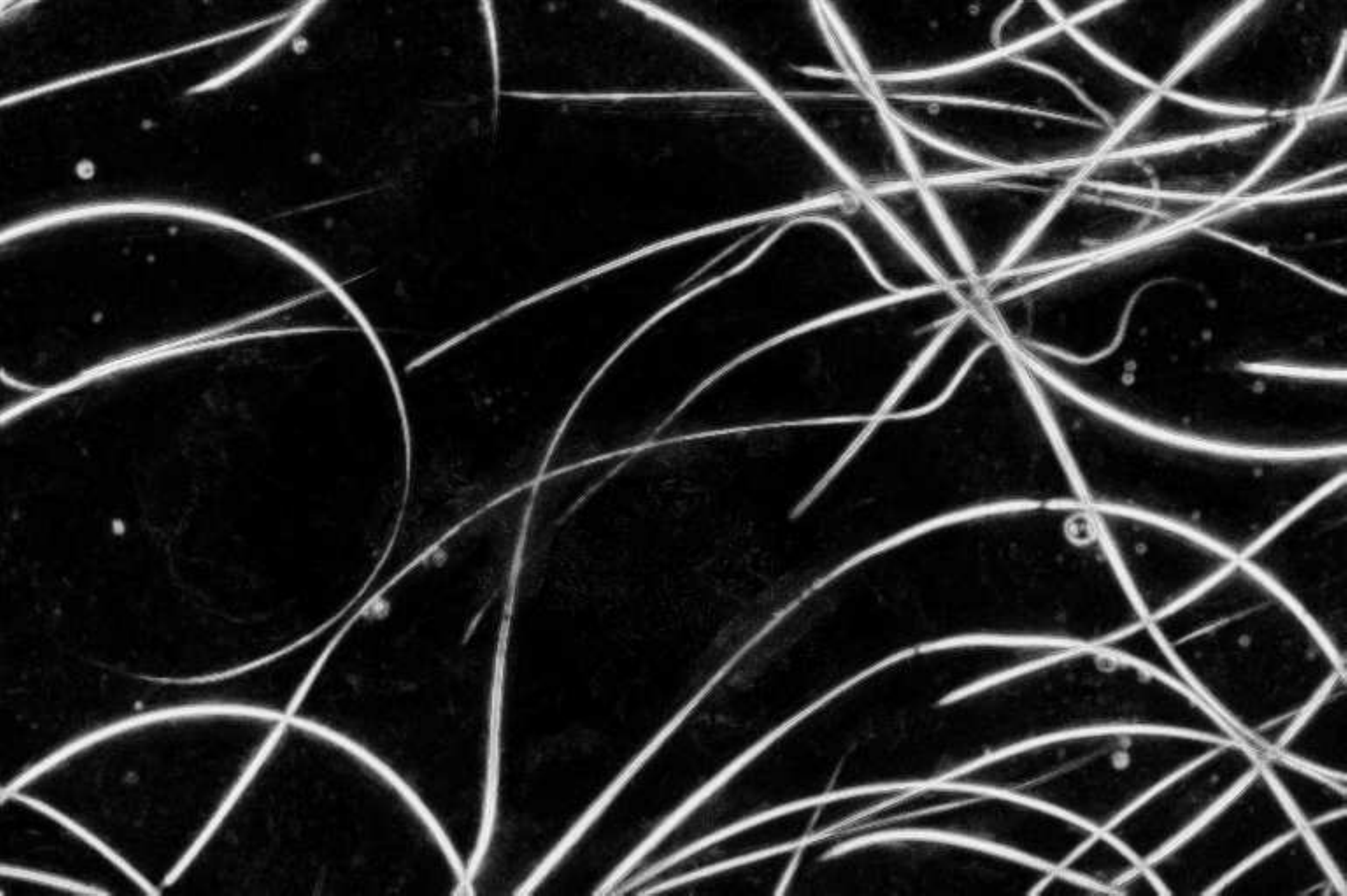}
		\includegraphics[width=0.2\columnwidth]{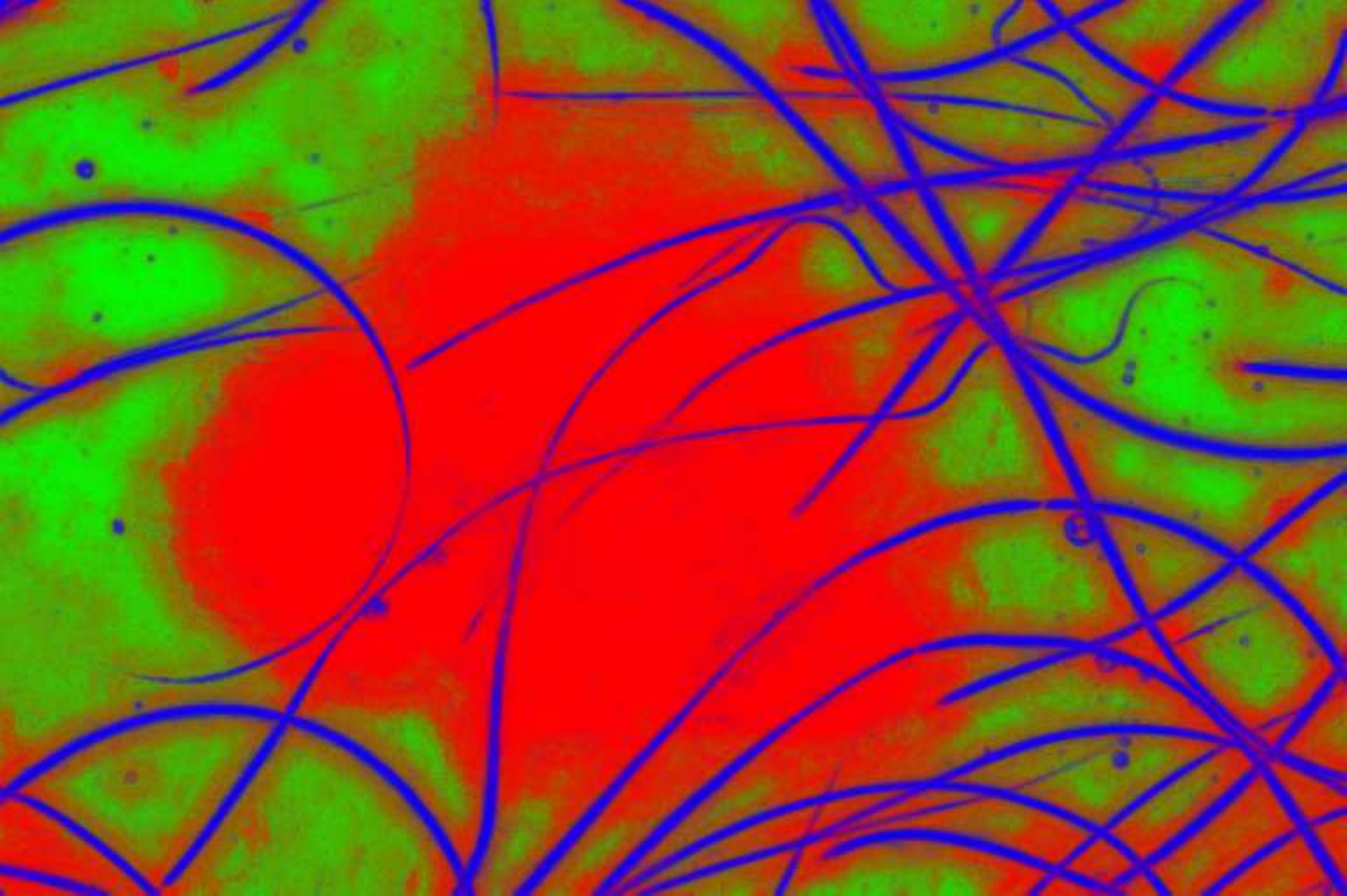}				
	}
	\caption{Example of processing in the fuzzy classification of pixels of ``lesion'', ``skin'' and ``other'' types. The first image is the original ${I}$. The following three are the gray probability images ${I_{lesion}}$, ${I_{skin}}$ and ${I_{other}}$. The fifth is the color image ${I_{lesion\_skin\_other}}$.}
	\label{fig_lesion_skin_other}
\end{figure}

These probability images show the pixel probabilities in graphic format, also taking advantage of the fact that as there are three images, a color image ${I_{lesion\_skin\_other}}$ can be built by assigning the ${I_{lesion}}$ to the red channel, ${I_{skin}}$ to the green channel and ${I_{other}}$ to the blue channel. This new image is equivalent to the three probability images and enables the result of the fuzzy classification of pixels to be seen much more clearly in graphic format, all of which is shown in Fig. \ref{fig_lesion_skin_other}.

\subsection{Module 3. Segmentation based on thresholding}
\label{Segmentation based on thresholding}


The authors were able to ascertain that histogram thresholding methods work for segmentation purposes very well in a high percentage of images. In fact, this had been reported in previous state-of-the-art studies. However, the major problem in most cases was that disturbing artifacts spoiled the search for the threshold, which will be addressed here. 

\begin{figure}[hbt]
	\centering
	\centerline{
		\includegraphics[width=1\columnwidth]{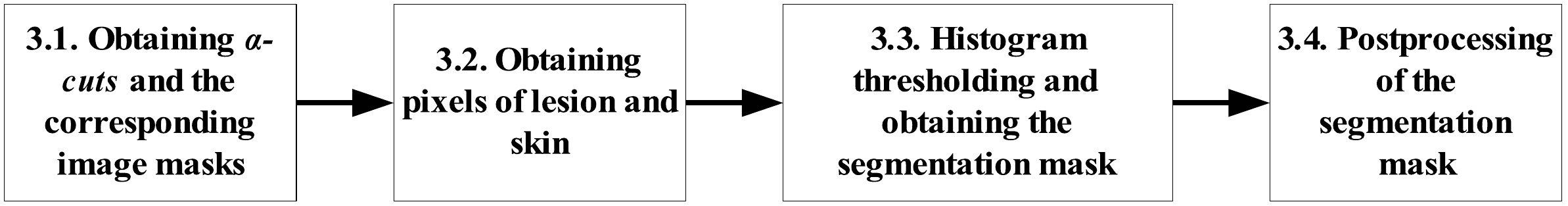}
	}
	\caption{Phases of the segmentation based on thresholding module.}
	\label{fig_segmentacion_thresholding}
\end{figure}

This module will consist of four phases, as can be seen in Fig. \ref{fig_segmentacion_thresholding}. Firstly, the ${\alpha-cuts}$ (which will be defined below) and their corresponding image masks can be obtained for different levels of probability from fuzzy sets and probability images. Secondly, different operations are undertaken using these masks in order to generate an image to which a histogram thresholding method can be applied. Thirdly, the histogram thresholding is applied, which enables the segmentation mask corresponding to the lesion to be obtained. Lastly, this mask is post-processed.

\subsubsection{3.1. Obtaining ${\alpha-cuts}$ and the corresponding image masks}
\label{Obtaining alpha-cuts y and the corresponding image masks}


The ${\alpha-cuts}$ and their corresponding image masks can be obtained for different levels of probability from fuzzy sets and probability images. ${\alpha-cut}$ ${[\mu]_{\alpha}\subset{X}}$, is defined given a fuzzy set ${\mu}$ of a set ${X}$ and a probability value  ${0\le\alpha\le1}$, as ${[\mu]_{\alpha}=\{(x,y)\in X:\mu(x,y)\le\alpha\}}$. From this, given the three probability valued ${\alpha_{lesion}}$, ${\alpha_{skin}}$ and ${\alpha_{other}}$, the following masks are obtained: ${BW^{\alpha_{lesion}}_{lesion}=\{(x,y)\in X:I_{lesion}(x,y)\le255.\alpha_{lesion}\}}$, extracted from  ${I_{lesion}}$ and corresponding to ${[\mu_{lesion}]_{\alpha_{lesion}}}$, ${BW^{\alpha_{skin}}_{skin}=\{(x,y)\in}$ 
${X:I_{skin}(x,y)\le{255.\alpha_{skin}}\}}$, extrac- ted from  ${I_{skin}}$ and corresponding to ${[\mu_{skin}]_{\alpha_{skin}}}$, and ${BW^{\alpha_{other}}_{other}=\{(x,y)\in}$ 
${X:I_{other}(x,y)\le{255.\alpha_{other}}\}}$, extracted from  ${I_{other}}$ and corresponding to ${[\mu_{other}]_{\alpha_{other}}}$.


These masks will be used later on in different phases of the module.

\subsubsection{3.2. Obtaining pixels of lesion and skin}
\label{Obtaining pixels of lesion and skin}


Disturbing artifacts are detected and lesion and skin pixels are obtained in this phase with a view to the histogram thresholding carried out in the following phase. This of course means that the mask obtained will have the maximum possible number of lesion and skin pixels, but above all and most importantly, this is to ensure that the minimum possible number of erroneous pixel colors –-corresponding to disturbing artifacts-– are included. The reason for this is obvious, that from among all the lesion and skin pixels, thresholding should not be too affected if part of the pixels are not taken into account. However, the presence of colors corresponding to erroneous pixels may give rise to erroneous behavior of the thresholding, especially in a certain type of image, and so this idea is important in helping to understand how the method works in this phase. 


To obtain the mask of lesion and skin pixels, using a threshold value ${THR\_OTHER=0.5}$ established empirically, the mask ${BW_{disturbingartifacts}=BW^{THR\_OTHER}_{other}}$ is the first approximation to the mask of disturbing artifact pixels and the mask ${BW_{lesion\_skin}=\complement{(BW_{disturbingartifacts})}}$ is the first approximation to the mask of lesion and skin pixels. An erosion operation is then carried out on this last-mentioned mask, the largest 8-connected component taken and the holes covered, then being eroded again. All this can be seen in graphic format in Fig. \ref{fig_otsu}.


The resulting mask ${BW_{lesion\_skin}}$ have lesion and skin pixels with high probability, although a modification will be made to the image on which this method is going to be applied in order to ensure greater certainty that the histogram thresholding method will work well. Using a threshold value ${THR\_SKIN=0.5}$ established empirically, the mask  ${BW_{skin}=BW^{THR\_SKIN}_{skin}}$ would be an approximation to the mask with skin pixels. Skin color is calculated from this mask ${BW_{skin}}$ as the median of the pixel colors and a blurred image ${I_{blurred}}$ is subsequently created using the values of the original image ${I}$, onto which this skin color is painted in ${BW_{disturbingartifacts}}$ pixels and median filter is then applied to it. Once this phase has been completed, the image ${I_{forthresholding}}$ is obtained from the original image ${I}$ by painting the colors of the ${I_{blurred}}$ pixels on the ${BW_{disturbingartifacts}}$ pixels. All this can be seen in graphic format in Fig. \ref{fig_otsu}.

\subsubsection{3.3. Histogram thresholding and obtaining the segmentation mask}
\label{Histogram thresholding and segmentation}

\begin{figure}[hbt]
	\centering
	\centerline{
		\includegraphics[width=0.25\columnwidth]{ISIC_0014876-eps-converted-to.pdf}
		\includegraphics[width=0.25\columnwidth]{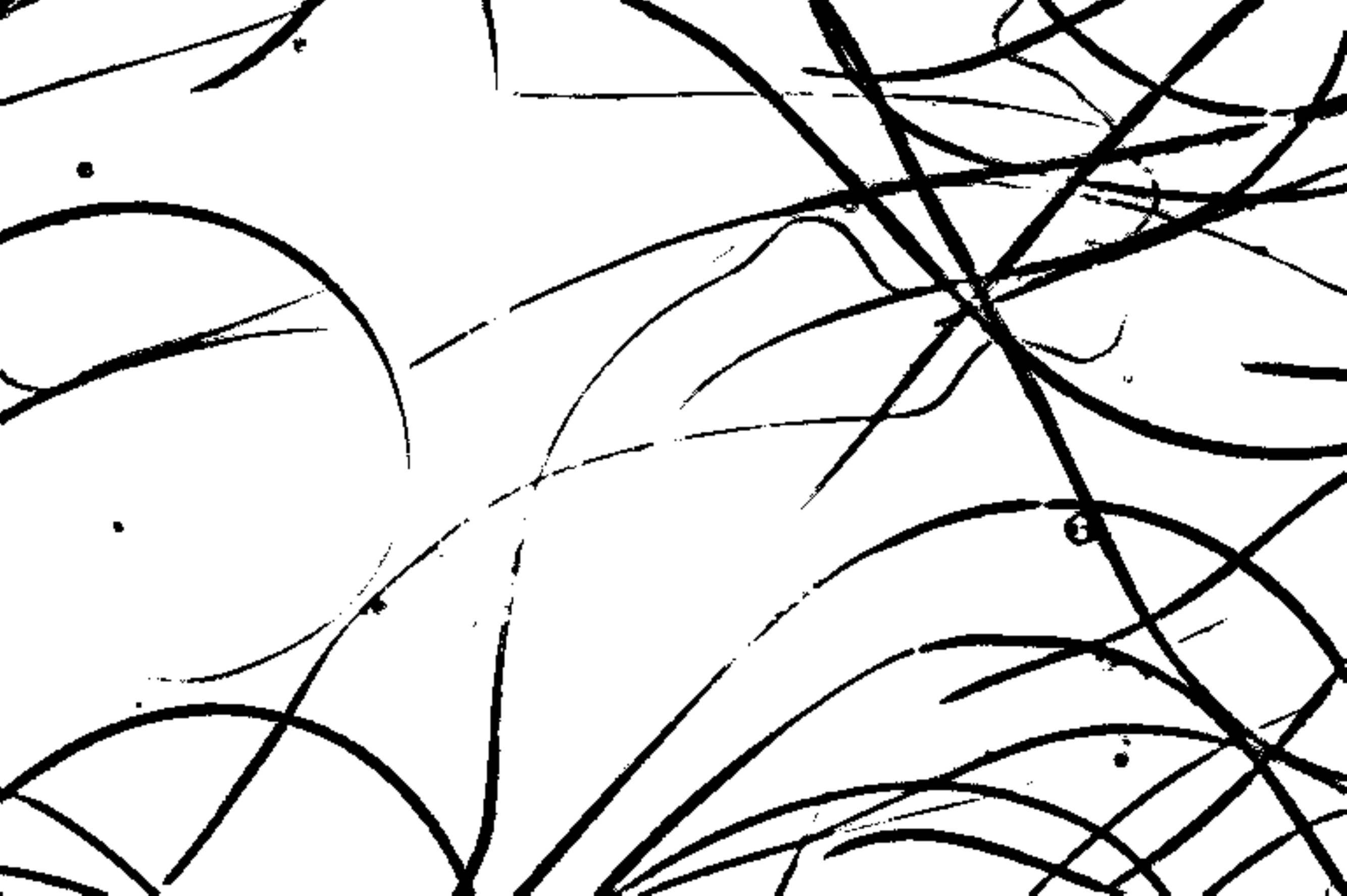}
		\includegraphics[width=0.25\columnwidth]{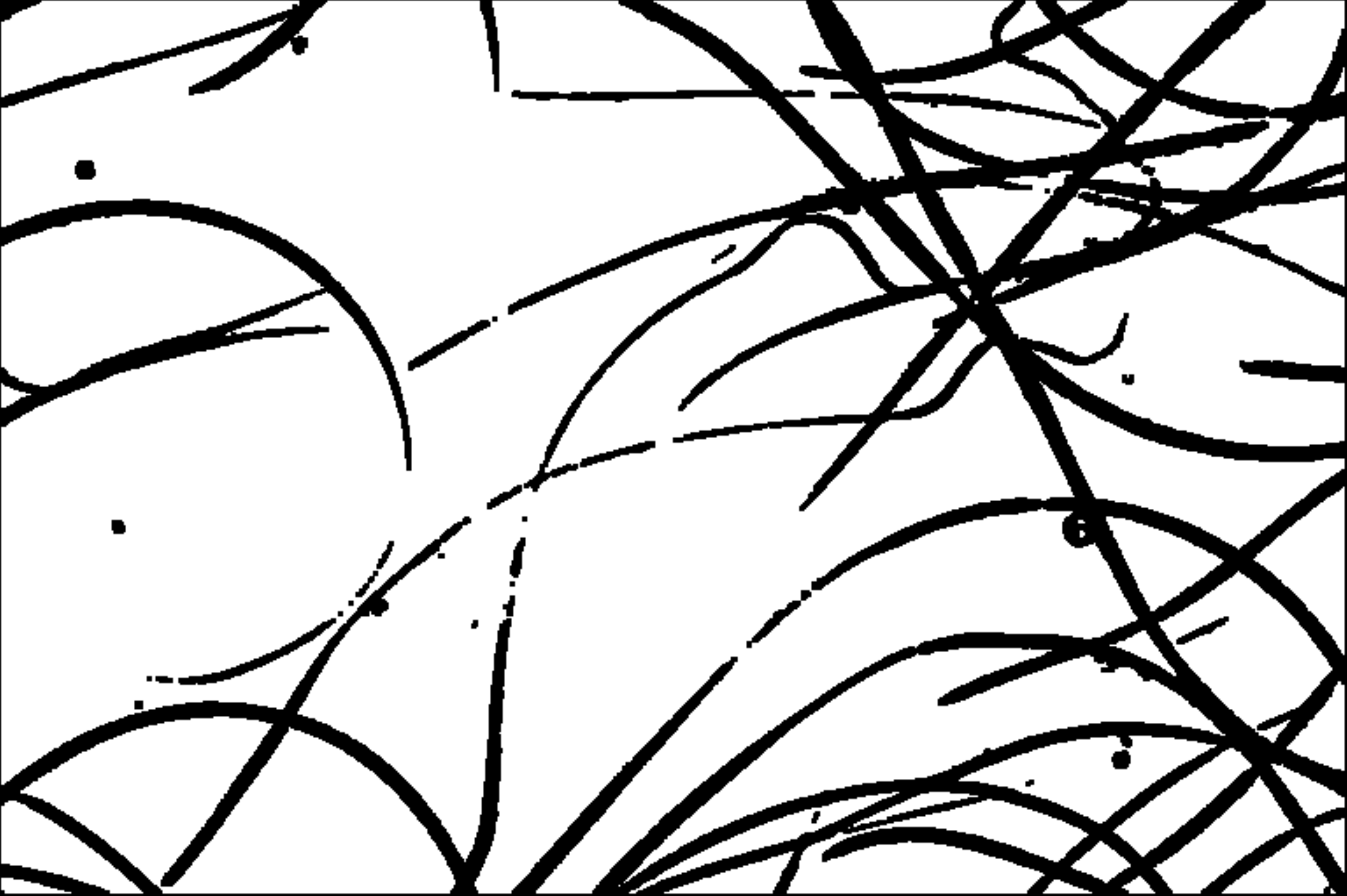}
		\includegraphics[width=0.25\columnwidth]{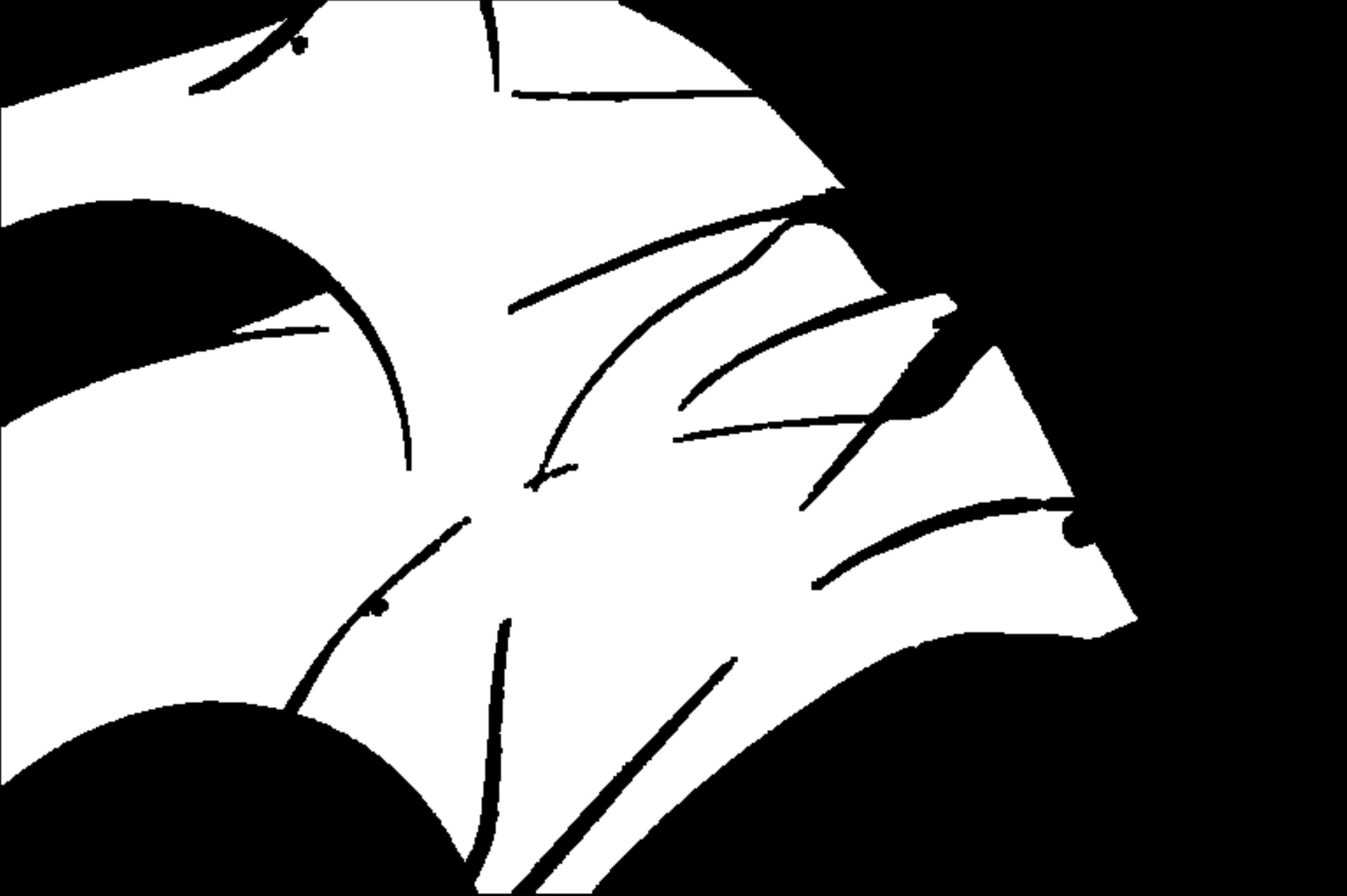}		
	}
	\centerline{
		\includegraphics[width=0.25\columnwidth]{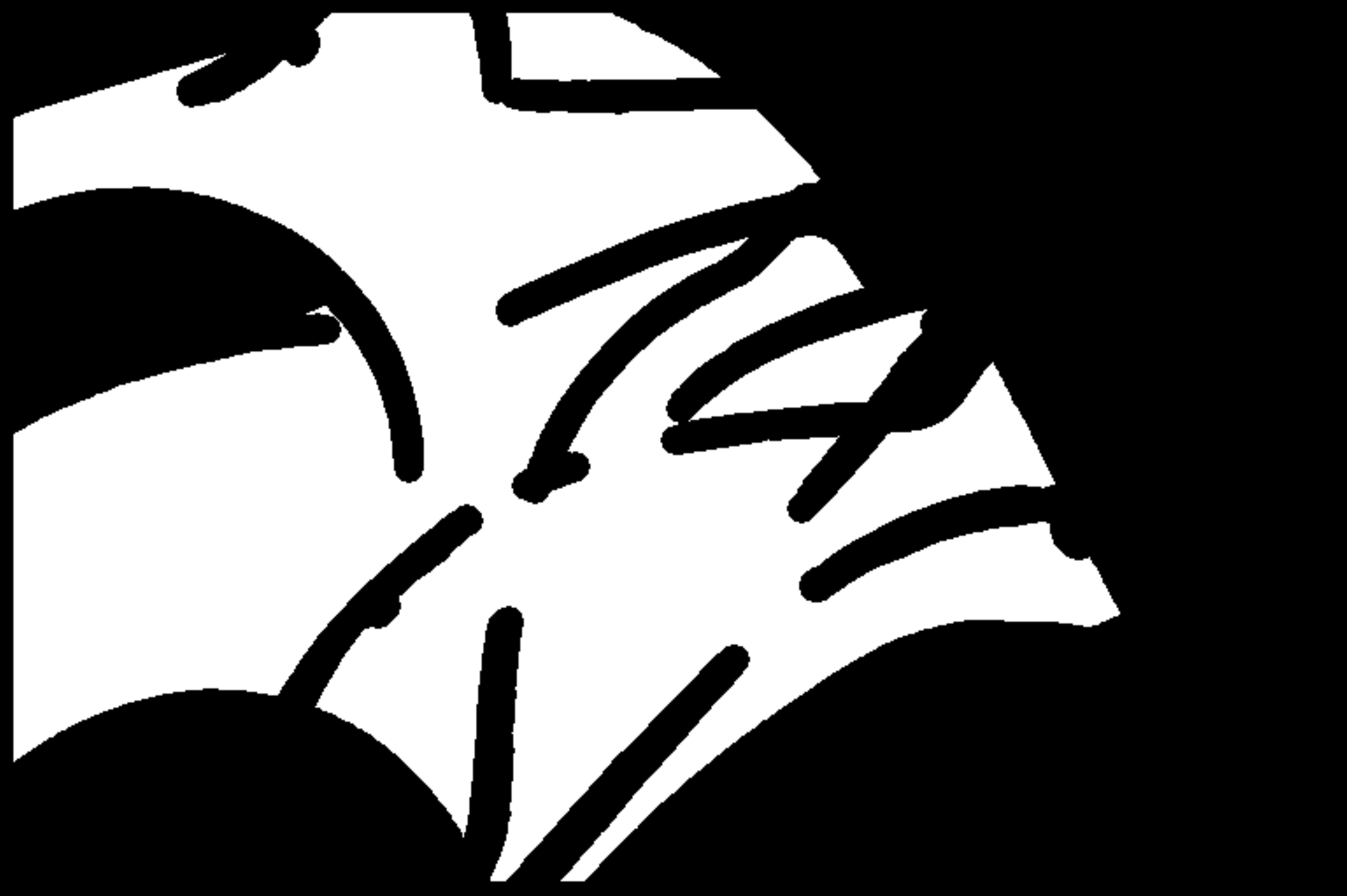}
		\includegraphics[width=0.25\columnwidth]{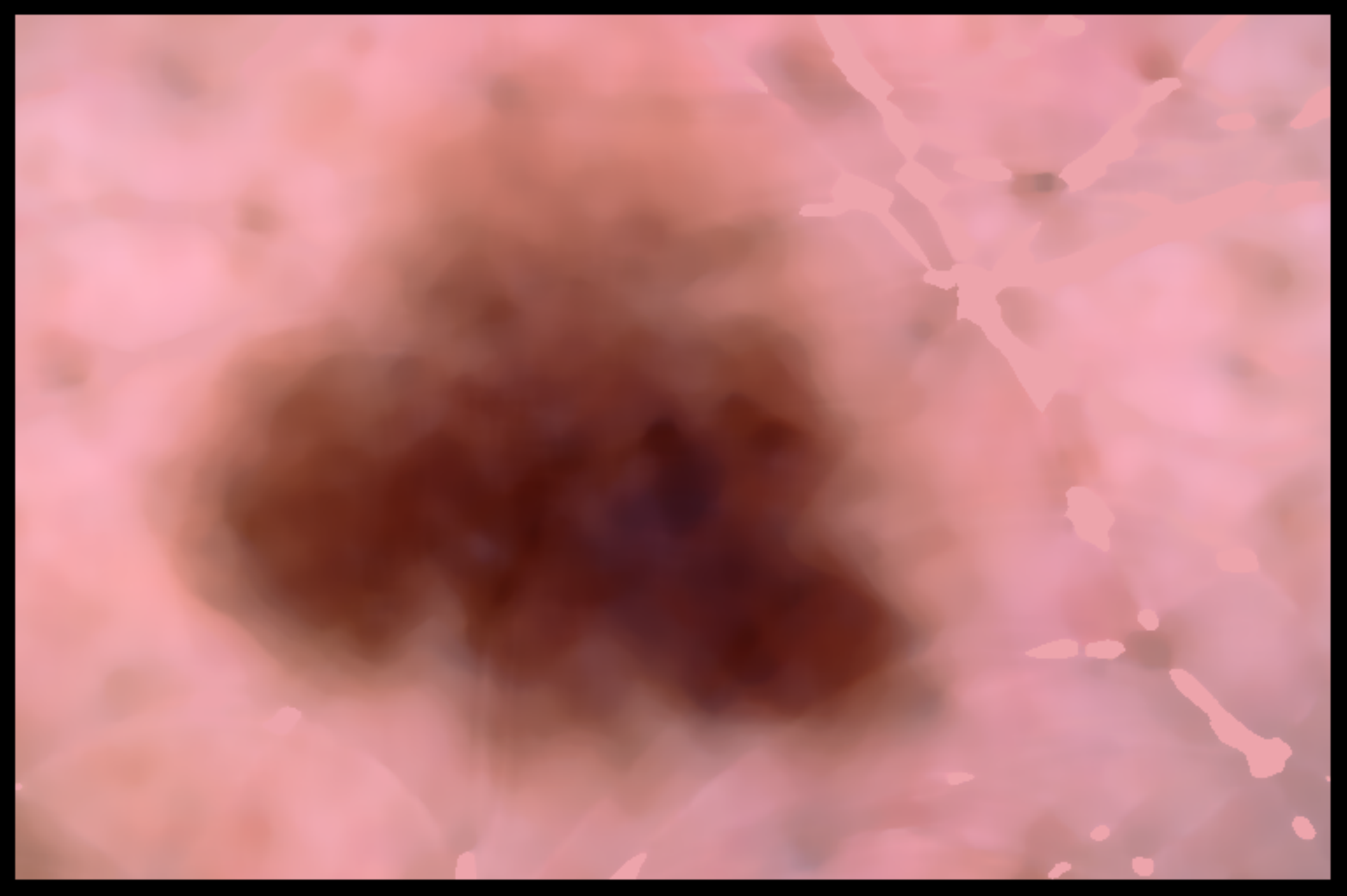}
		\includegraphics[width=0.25\columnwidth]{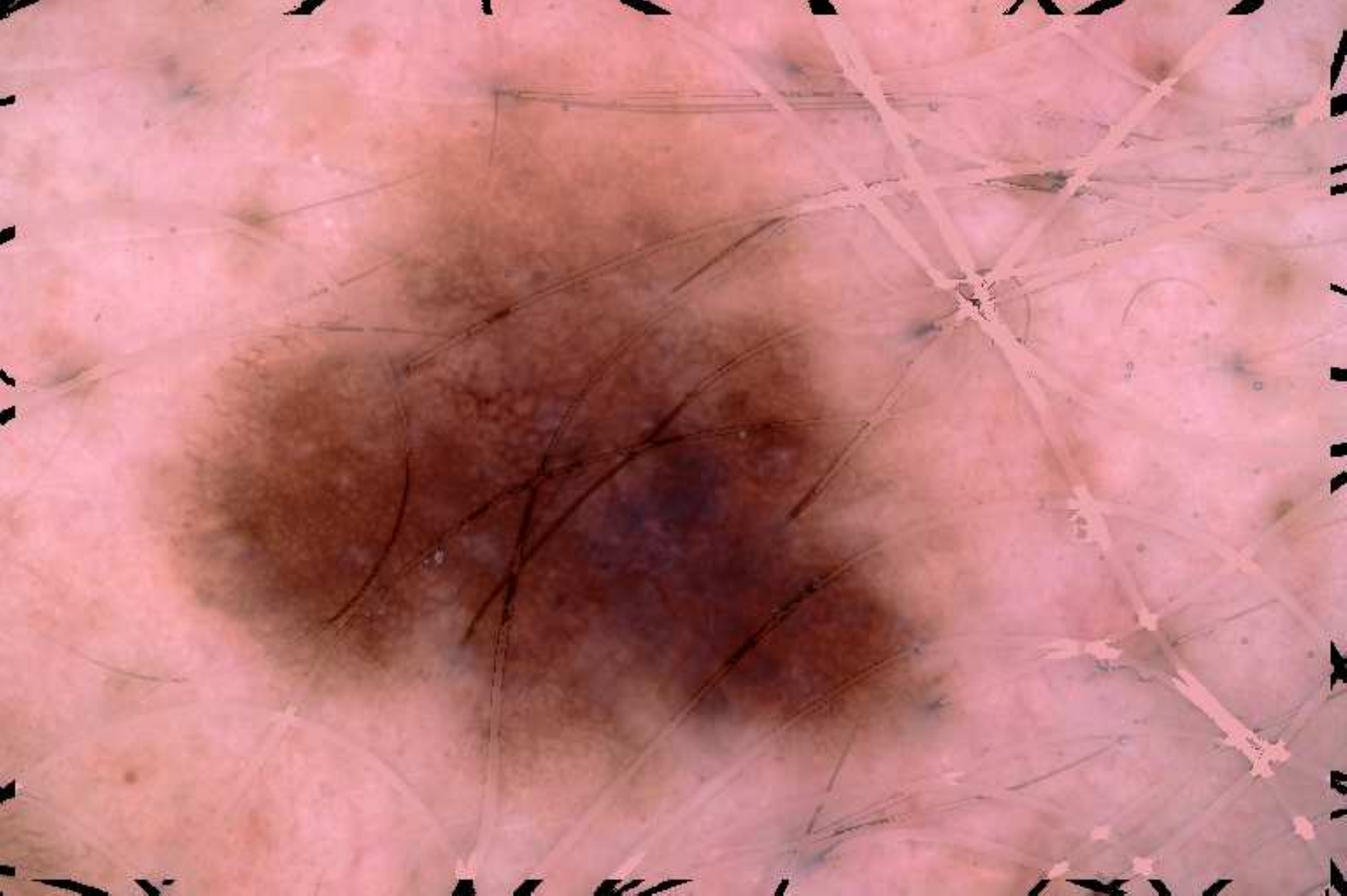}
		\includegraphics[width=0.25\columnwidth]{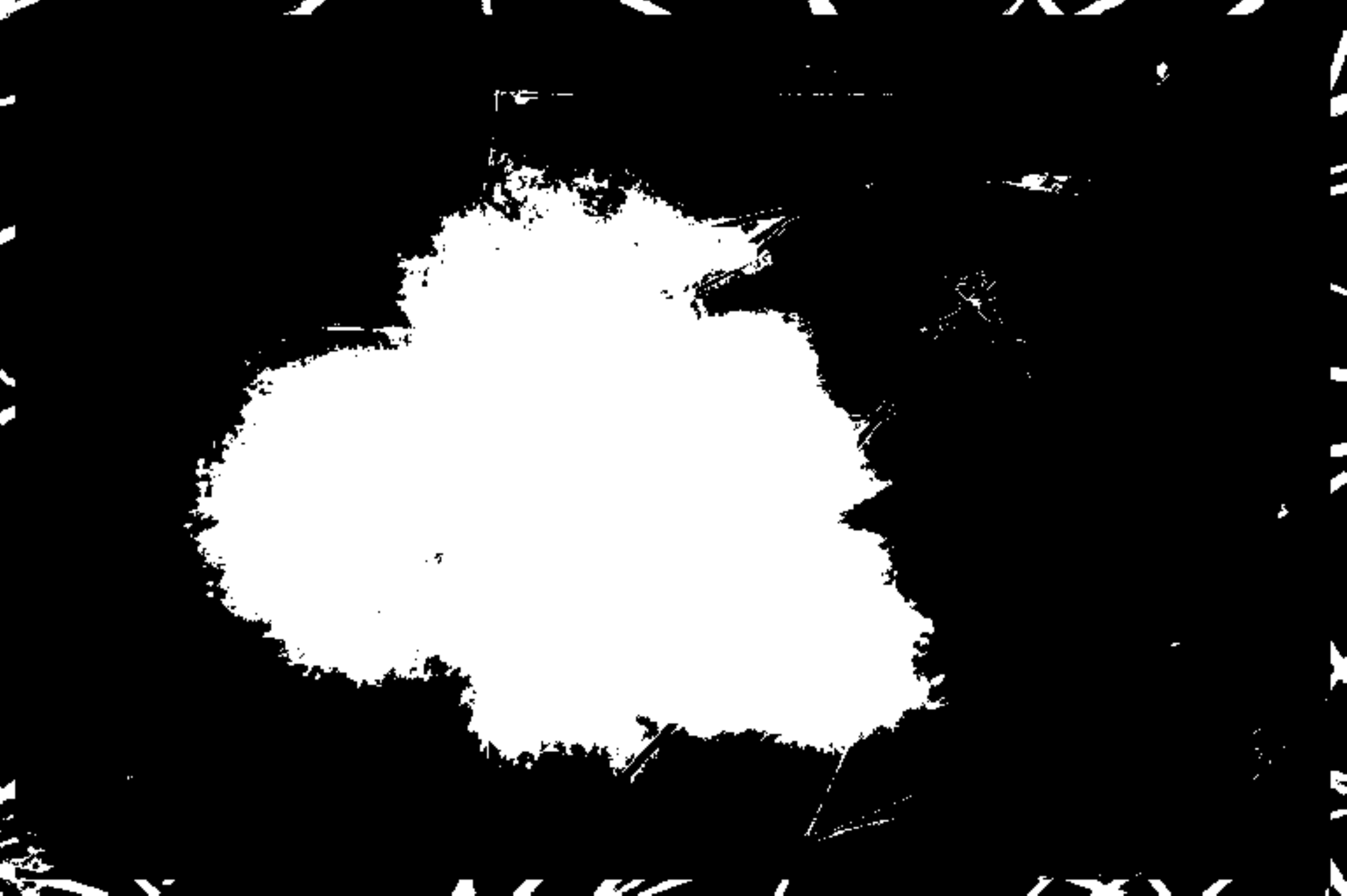}		
	}
	\caption{Some images of the process: 1. Original image; 2, 3, 4 and 5: ${BW_{lesion\_skin}}$ at different moments; 6: ${I_{blurred}}$ generated; 7: ${I_{forthresholding}}$; 8: ${BW_{maskoflesion}}$.}
	\label{fig_otsu}
\end{figure}

The histogram threshold value for the image ${I_{forthresholding}}$ is calculated in the mask ${BW_{lesion\_skin}}$ and the segmentation mask ${BW_{maskoflesion}}$ is calculated. The Otsu method \cite{Otsu1975} is used on the blue channel, which is the histogram thresholding method that worked best among all those tried out and which has been previously used in many other methods \cite{Celebi2015}. This can be seen in graphic format in Fig. \ref{fig_otsu}.

\subsubsection{3.4. Postprocessing of the segmentation mask}
\label{Postprocessing of the segmentation mask}


Once ${BW_{maskoflesion}}$ has been obtained, the image then needs to be post-processed in order to resolve any problems there may be as follows: 1. To prevent hairs from cutting the mask; 2. To prevent the edges (both rectangular and above all circular), which are dark in many of the images, from remaining as part of the lesion mask; 3. To reduce any effects of disturbing artifacts as far as possible; 4. To set soft edges, to ensure there are not too many recesses or projections and that there is certain convexity in the resulting mask; 5.- As commented in \ref{High level view}, ensuring the most accurate possible segmentation, albeit always giving priority to sensitivity over specificity.


Firstly, to refine the mask ${BW_{maskoflesion}}$ in preventing hair from cutting it and dark edges from sticking to the lesion mark, the lesion pixels are divided into two separate parts as follows: ${BW_{interior}}$ and ${BW_{exterior}=\complement{(BW_{interior})}}$, using a radius ${radius=minimum(\frac{3}{8}.width,\frac{3}{8}.height)}$. Then the ${BW_{lesion\_skin}}$ is taken at its original value ${BW_{lesion\_skin}=\complement{(BW_{disturbingartifacts})}=}$\\
${\complement{(BW^{THR\_OTHER}_{other})}}$, which is dilated into ${BW_{interior}}$ and eroded into ${BW_{exterior}}$. Both intersect with ${BW_{maskoflesion}}$ and among the largest 8-connected components the most centered is taken and the holes are covered.

\begin{figure}[hbt]
	\centering
	\centerline{
		\includegraphics[width=0.25\columnwidth]{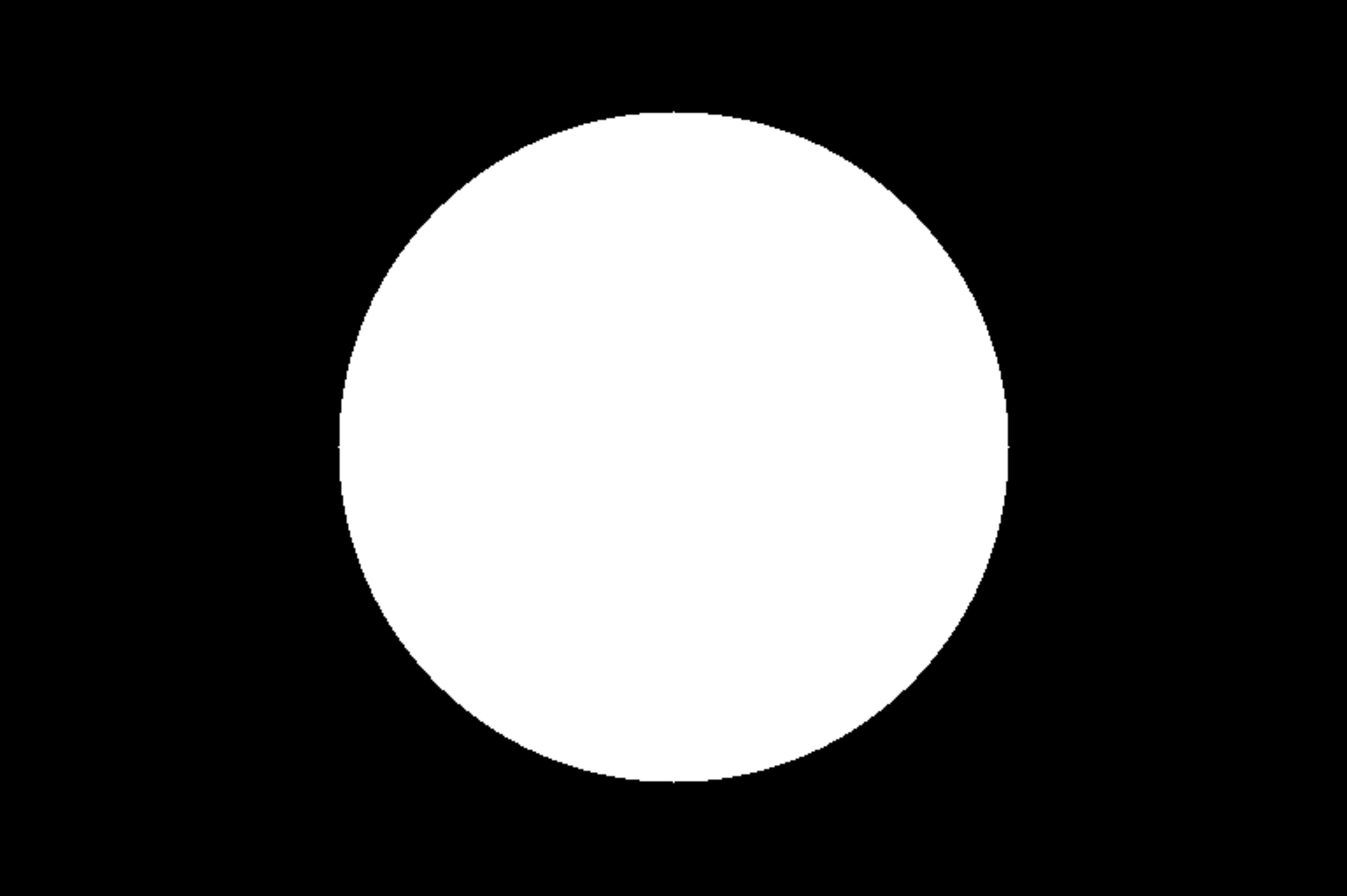}
		\includegraphics[width=0.25\columnwidth]{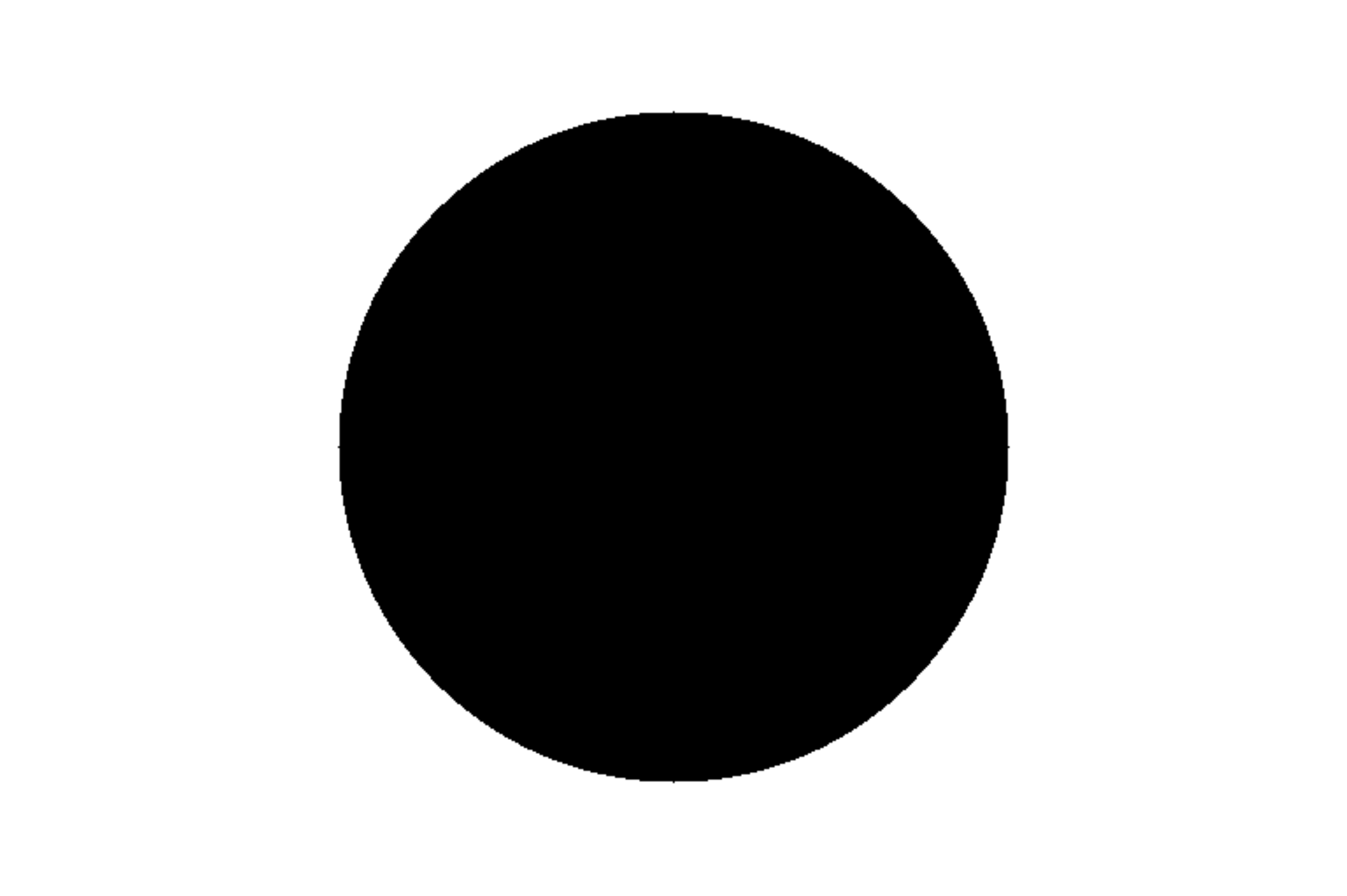}
		\includegraphics[width=0.25\columnwidth]{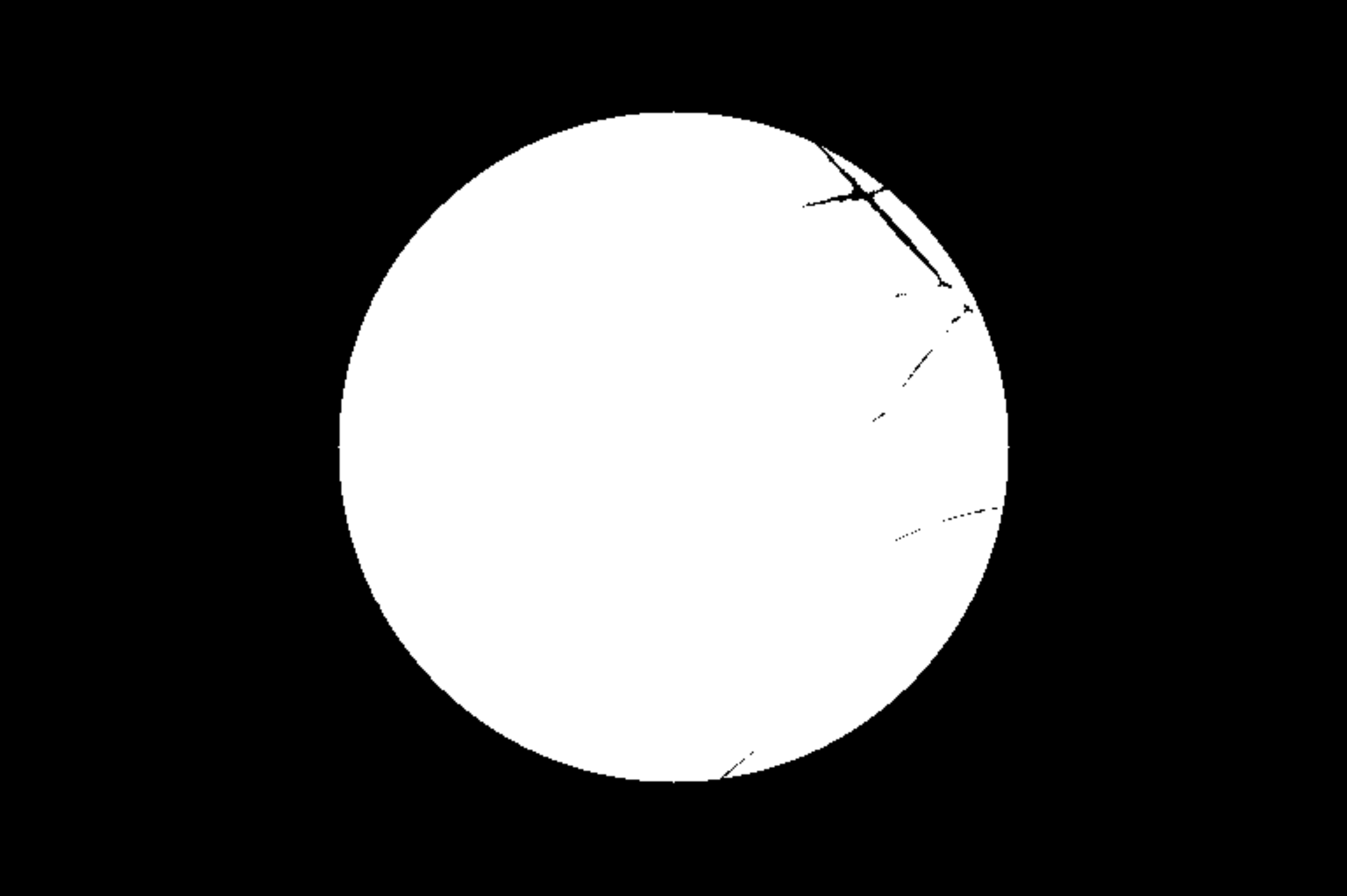}
		\includegraphics[width=0.25\columnwidth]{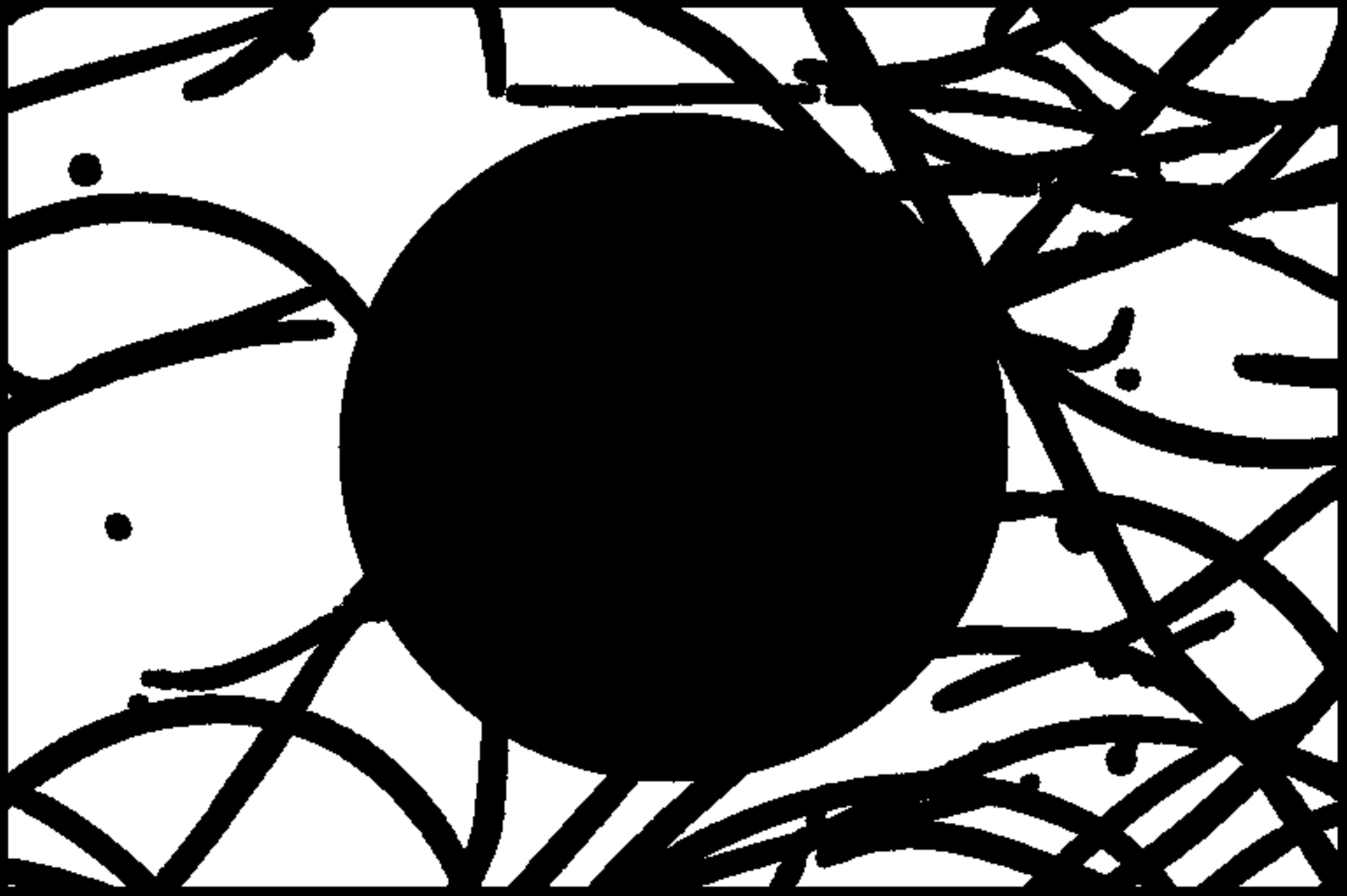}		
	}
	\centerline{
		\includegraphics[width=0.25\columnwidth]{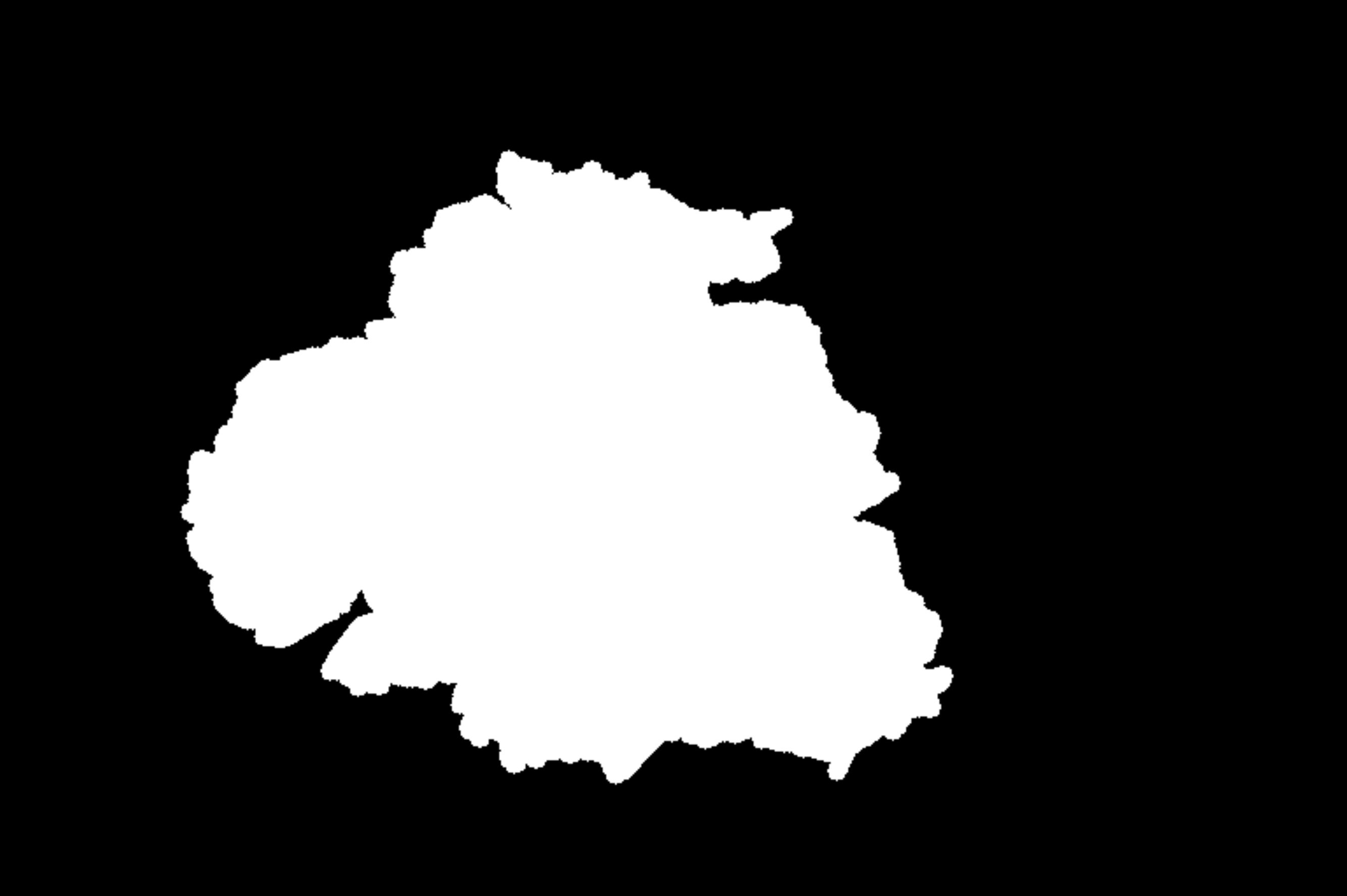}
		\includegraphics[width=0.25\columnwidth]{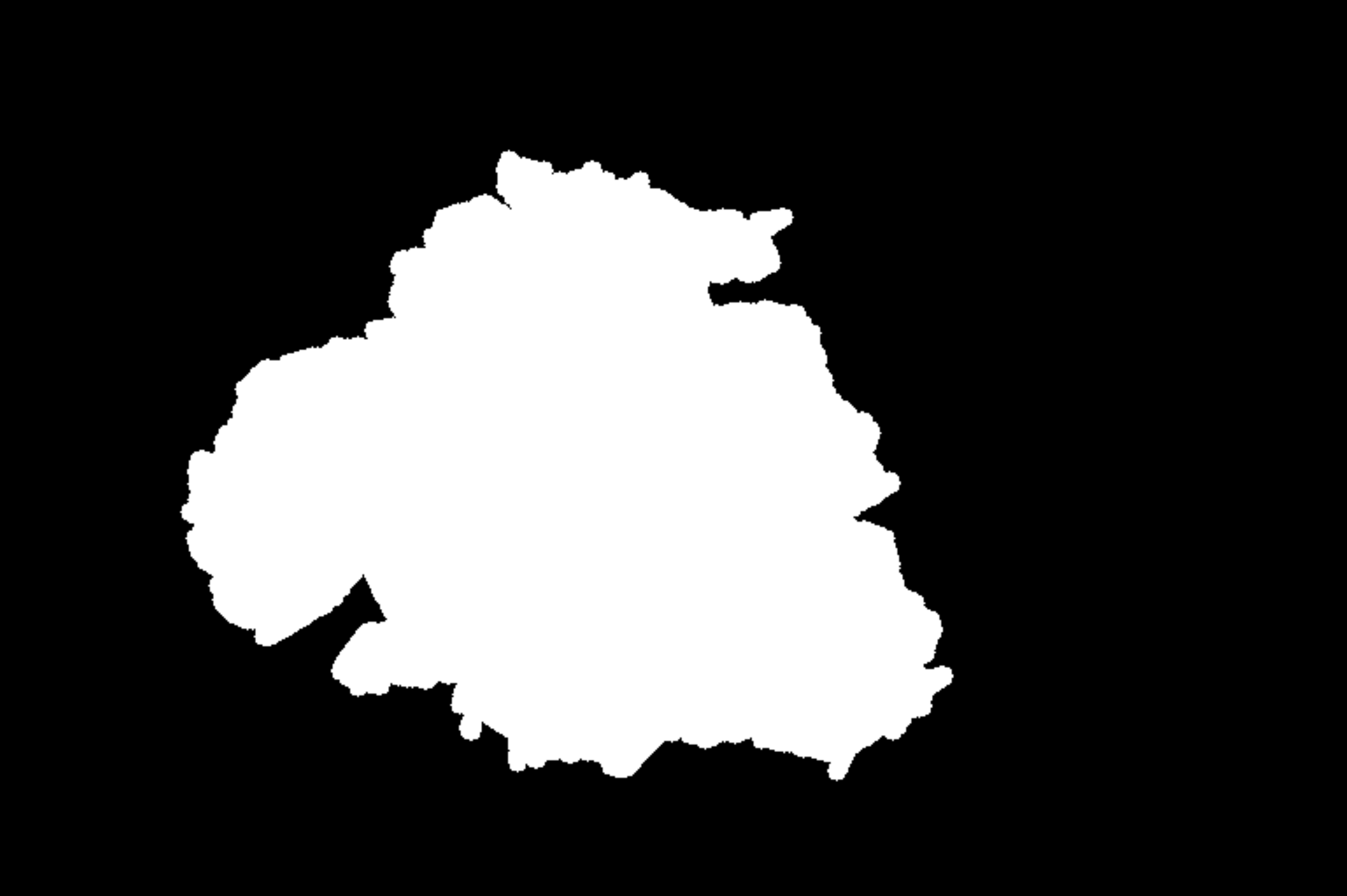}
		\includegraphics[width=0.25\columnwidth]{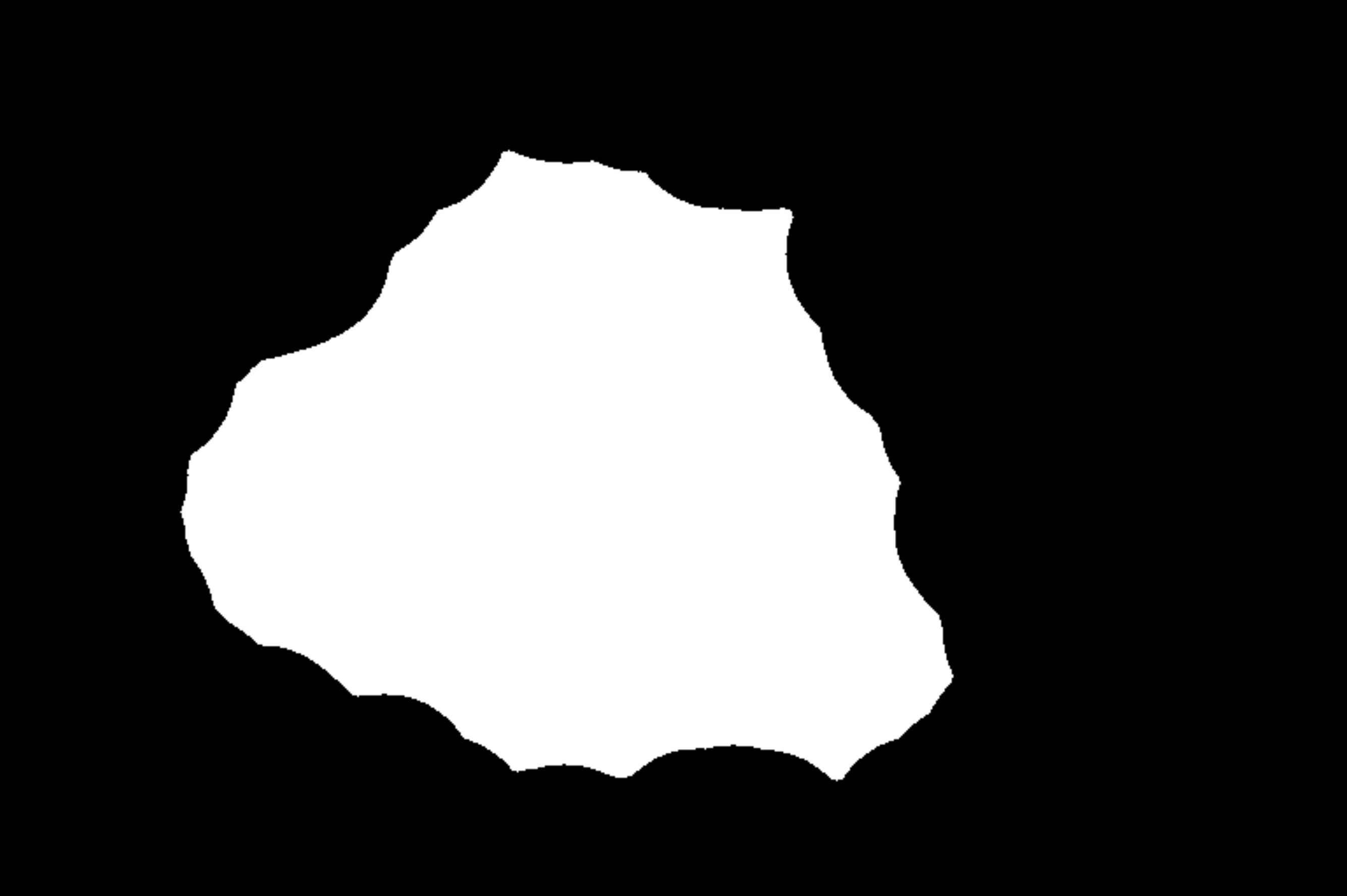}	
		\includegraphics[width=0.25\columnwidth]{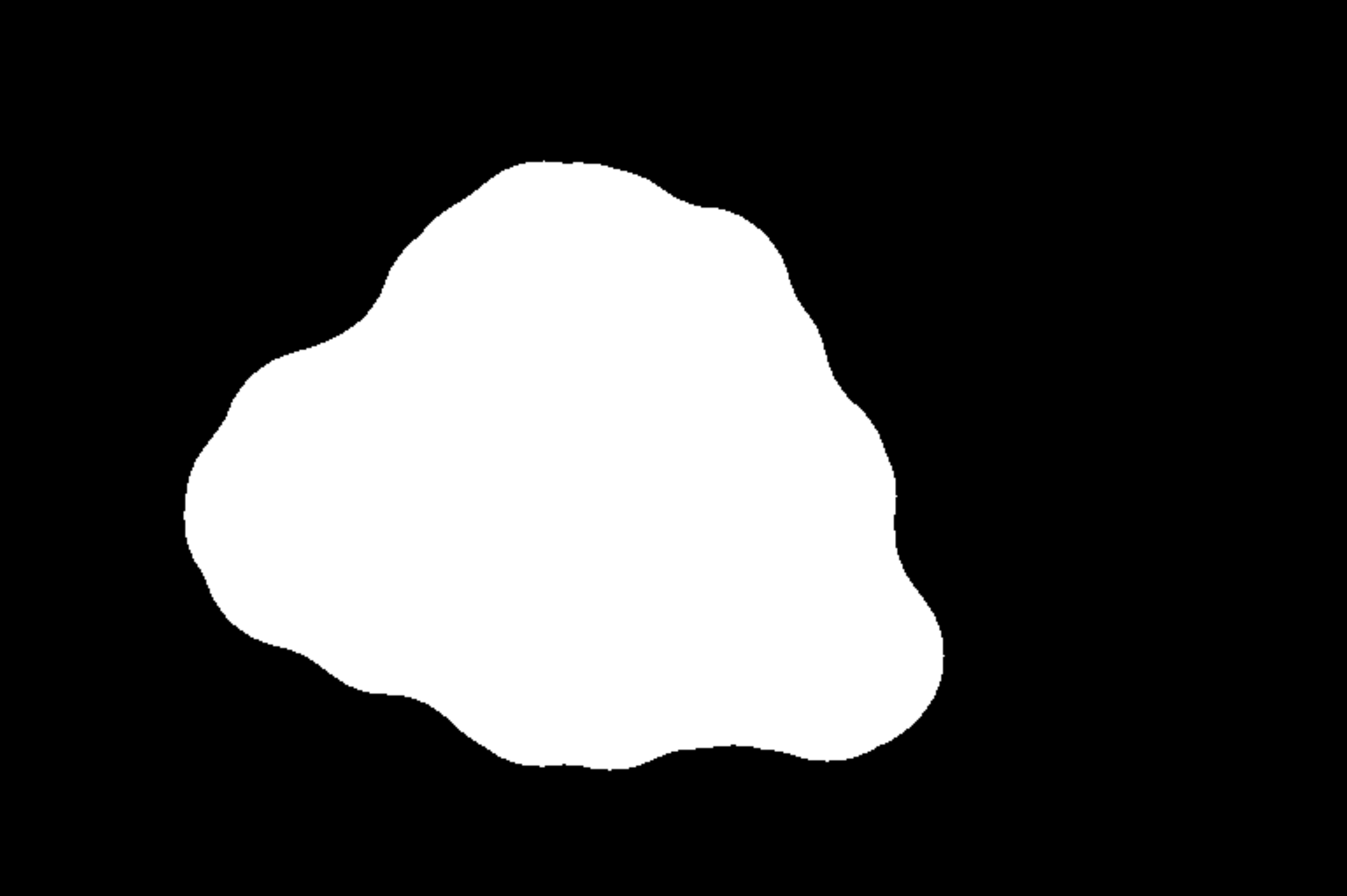}	
	}
	\caption{Some images from the process: 1: ${BW_{interior}}$; 2: ${BW_{exterior}}$; 3 and 4: ${BW_{lesion\_skin}}$ at its original value, dilated into  ${BW_{interior}}$ and eroded into ${BW_{exterior}}$, respectively. 5: ${BW_{maskoflesion}}$ after intersecting with the union of 3 and 4, taken among the largest 8-connected components the most centered, and covered the holes; 6, 7 and 8: ${BW_{maskoflesion}}$ following application of different morphological operations.}
	\label{fig_postprocessing}
\end{figure}  

Then different morphological operations (eroding and dilating) are then carried out in order --without breaking the mask (as it is very narrow in some parts)-- to reduce the effects of the disturbing artifacts, remove any projections, cover any gaps, soften the edges and ensure that there is certain convexity in the resulting mask.  All this can be seen in Fig. \ref{fig_postprocessing}.


\section{Results and discussion}
\label{Results and discussion}



\subsection{Fuzzy classification of pixels of type ``lesion'', ``skin'' and ``other''}
\label{Results Fuzzy classification of pixels of lesion, skin and other}

As mentioned in \ref{Obtaining the clasification model}, the purpose of the fuzzy detection of pixels is to obtain fuzzy membership rules for each of the pixels. The Weka implementation of Random Forest is used \cite{Breiman2001}, which is a classifier that provides such functionality (a fuzzy classifier) and, as shall be seen, obtains very good results in terms of reliability, as well as being very fast with low computational cost. This is essential in order to make the algorithm efficient, since each of the pixels of the different images must be computed. From a set of 30444 samples taken, 94.69\% accuracy and AUC of 0.992 were obtained using 10-fold cross-validation.

\subsection{Lesion Segmentation}
\label{Lesion Segmentation}


As referred to in \ref{Evaluation of the method}, the metrics defined in the 2016 \cite{Gutman2016} and 2017 \cite{ChallengeISBI2017} ISBI Challenges were used to evaluate the methods: Accuracy, Dice Coefficient, Jaccard Index, Sensitivity and Specificity. In the 2016 ISBI Challenge, the results 0.934, 0.869, 0.791, 0.870 and 0.978 respectively were obtained with a previous version of this method, and in the 2017 ISBI Challenge, the results 0.895, 0.750, 0.651, 0.894 and 0.918 respectively were obtained with the current improved version in the validation phase --not the final results--, using 150 images. The difficulty attached to the test data set obviously also needs to be taken into account in order to assess these results. In any case, it should be pointed out that, as referred to in \ref{High level view}, an attempt was made in this improved version of the method to give priority to sensitivity, even if this meant penalizing other indexes, which ensures that this method is very robust against other images from different data sets. In fact, the method has also been tested using other data sets, obtaining good results.


\ifCLASSOPTIONcaptionsoff
  \newpage
\fi



\bibliographystyle{IEEEtran}
\bibliography{references}

%

\end{document}